\documentclass[a4paper,11pt,twocolumn,twoside]{article}
\usepackage[final]{graphicx}
\usepackage{sepln_en}
\usepackage{fullname}
\usepackage[utf8]{inputenc}

\input epsf

\newcommand{\hum}{\textsc{Hum}}
\newcommand{\gen}{\textsc{Gen}}
\newcommand{\set}[1]{\{#1\}}

\usepackage{amssymb}
\usepackage{amsfonts}
\usepackage{amsmath}
\usepackage{colortbl}
\usepackage{comment}
\usepackage[
left = ``,%
right = ''%
]{dirtytalk}
\usepackage{multirow}
\usepackage{subcaption}
\usepackage{xcolor}
\usepackage{url}

\title{Overview of AuTexTification at IberLEF 2023: Detection and Attribution of Machine-Generated Text in Multiple Domains}
\author {\textbf{Areg Mikael Sarvazyan},$^1$ \textbf{José Ángel González},$^1$ \textbf{Marc Franco-Salvador},$^1$\\ \textbf{Francisco Rangel},$^1$ \textbf{Berta Chulvi},$^2$ \textbf{Paolo Rosso$^2$}\\
$^1$Symanto Research, Valencia, Spain\\
$^2$Universitat Politècnica de València, Valencia, Spain\\
\set{areg.sarvazyan, jose.gonzalez, marc.franco, francisco.rangel}@symanto.com\\
berta.chulvi@upv.es, prosso@dsic.upv.es}

\seplntranstitle{Resumen de AuTexTification en IberLEF 2023:
Detección y Atribución de Texto Generado Automáticamente en Múltiples Dominios}

\seplnclave{Texto Generado por Máquina, Modelos de Lenguaje Masivos, Generalización, AuTexTification.}

\seplnresumen{Este artículo presenta un resumen de la tarea AuTexTification como parte del workshop IberLEF 2023 sobre el Iberian Languages Evaluation Forum, en el marco de la conferencia SEPLN 2023. AuTexTification consta de dos subtareas: en la Subtarea 1, los participantes tuvieron que determinar si un texto fue escrito por un humano o generado por un modelo de lenguaje masivo. Para la Subtarea 2, los participantes debían atribuir un texto generado automáticamente a uno de seis modelos de generación de texto diferentes. El conjunto de datos AuTexTification contiene más de 160.000 textos en dos idiomas (inglés y español) y cinco dominios (tweets, reseñas, noticias, legislación y artículos instructivos). Un total de 114 equipos se inscribieron para participar, de los cuales 36 enviaron 175 resultados y 20 de ellos enviaron artículos. En este artículo, presentamos el conjunto de datos y la tarea AuTexTification, los sistemas enviados por los participantes y sus resultados.}

\seplnkey{Machine-Generated Text, Large Language Models, Generalization, AuTexTification.}

\seplnabstract{This paper presents the overview of the AuTexTification shared task as part
of the IberLEF 2023 Workshop in Iberian Languages Evaluation Forum, within the framework of the SEPLN 2023 conference. AuTexTification consists of two subtasks: for Subtask 1, participants had to determine whether a text is human-authored or has been generated by a large language model.
For Subtask 2, participants had to attribute a machine-generated text to one of six different text generation models. 
Our AuTexTification 2023 dataset contains more than 160.000 texts across two languages (English and Spanish) and five domains (tweets, reviews, news, legal, and how-to articles). A total of 114 teams signed up to participate, of which 36 sent 175 runs, and 20 of them sent their working notes. In this overview, we present the AuTexTification dataset and task, the submitted participating systems, and the results.}

\firstpageno{1}

\begin{document}

\setlength\titlebox{21cm} 

\label{firstpage} \maketitle

\section{Introduction}
Current developments in Large Language Models (LLMs) have strongly improved the quality of Machine-Generated Text (MGT).
Their latest surge in popularity through services such as ChatGPT,\footnote{\url{https://tinyurl.com/reuters-chatgpt}} and large-scale democratization efforts to broaden the public's access to large models \cite{scao2022bloom,touvron2023llama,wolf2020transformers,seger2023democratising}, have made it easier for non-technical people to interact with and use these models for various interesting applications \cite{eloundou2023gpts,liu2023summary}.

\begin{table}[t]
\centering
\resizebox{\linewidth}{!}{%
\begin{tabular}{c|cc}
\multicolumn{1}{l|}{} & \multicolumn{2}{c}{\textbf{Model adaptation}} \\ \hline
\textbf{\begin{tabular}[c]{@{}c@{}}Human\\ Mod.\end{tabular}} & \textbf{Pre-trained} & \textbf{Fine-tuned} \\ \hline
No & \cellcolor{green!25}\begin{tabular}[c]{@{}c@{}} Full accessibility\\ Few comp. resources\\ Massive scale\\ Medium quality\end{tabular} & \begin{tabular}[c]{@{}c@{}}Technical accessibility\\ Large comp. resources\\ Massive scale\\ High quality\end{tabular} \\ \hline
Yes & \begin{tabular}[c]{@{}c@{}}High accessibility\\ Few comp. \& human resources\\ Small scale\\ High quality\end{tabular} & \begin{tabular}[c]{@{}c@{}}Technical accessibility\\ Large comp. \& human resources\\ Small scale\\ High quality\end{tabular}
\end{tabular}}
\caption{\label{tab:types-of-mgt}Types of MGT. The AuTexTification 2023 Shared Task focuses on generations from pre-trained models with no human modification. We cover the most accessible approach, involving little computational and human resources and can be used massively.}
\end{table}

However, these advances have also lowered the barrier of entry for users to generate high-quality, multi-style and multi-domain text in a massive scale. This means that motivated malicious users could easily generate massive quantities of text without the need of large computational resources, technical knowledge, or human intervention (see Table \ref{tab:types-of-mgt}). Supporting this concern, recent research suggest that disinformation generated with state-of-the-art LLMs is more credible than the one generated by humans \cite{Spitale23}, thus showing the difficulty for humans to distinguish between MGT and human-authored text.

As expected, the aforementioned advancements have also promoted discussions in ethical AI \cite{widder2022limits} as well as model, data and training regulations,\footnote{European Commission, Proposal for a Regulation of the European Parliament \url{https://tinyurl.com/EURAIAct}} and new licenses \cite{benjamin2019towards,contractor2022behavioral}.
Content moderation due to AI democratization, and the need for regulations, are strong motivators for researchers to ensure a responsible use of LLMs and their generations.
A promising research line to carry this out involves identifying MGT, while also attributing it to specific text generation models to learn about the specific actors behind an MGT from a forensics viewpoint.

There have been many efforts to detect MGT, including zero-shot approaches \cite{mitchell2023detectgpt,zellers2019defending}, supervised systems \cite{ippolito2020automatic,uchendu2020authorship,maronikolakis2021identifying}, and evaluation campaigns \cite{kashnitsky2022overview,shamardina2022findings}.
While it has been found that in-domain MGT detection with supervised approaches is easy \cite{bakhtin2019real}, most of the works often overlooked that MGT detection systems would be applied to a broad variety of domains, writing styles, and generation models.
Therefore, there is a need to evaluate the generalization of MGT detectors through a more realistic lens. In this regard, some works have studied generalization across model families and scales \cite{clef_autextification}, however, the generalization to new domains is still underexplored. 



In this context, we present the AuTexTification (\textbf{Au}tomated \textbf{Tex}t Iden\textbf{Tification}) task. This shared task is proposed to study: (i) the automatic detection of MGT, (ii) the generalization capabilities of MGT detectors to new domains, and (iii) the feasibility of fine-grained MGT attribution to one of many generation models.
Furthermore, we automatically collect a multi-domain annotated dataset of human-authored text and MGT generated by various LLMs, which is a valuable resource for exploratory linguistic analysis of machine-generated and human-authored texts.
To our knowledge, AuTexTification is the first shared task to study both MGT detection and attribution in a multi-domain setting for English and Spanish, while also focusing on generalization of MGT detectors to new domains. 



\section{Task Description}
The AuTexTification 2023 Shared Task includes two subtasks in English and Spanish in five different domains.

\paragraph{Subtask 1: MGT Detection.} This subtask consists in distinguishing between human and generated text. It is framed as a binary classification task of human text (\hum) and MGT (\gen), where text from three domains is included in the training set, and submissions are evaluated in two unseen ones.
This way, we aim to study the MGT detectors' cross-domain generalization capabilities.

\paragraph{Subtask 2: MGT Attribution.} In this subtask, participants must attribute MGT to the model that generated it, out of six models.
Thus, Subtask 2 is framed as a six-class classification task, where we strive to study the feasibility of fine-grained attribution. 
Differently to Subtask 1, the training and test splits include all five domains.


\section{Dataset}
The AuTexTification dataset consists of texts written by humans and LLMs in five domains: tweets, reviews, how-to articles, news and legal documents. 
These domains were chosen to encompass a range of writing styles, from more structured and formal to less structured and more informal. 
We collected human texts from publicly available datasets, namely: \textit{MultiEURLEX}~\cite{chalkidis-etal-2021-multieurlex}, \textit{XSUM}~\cite{Narayan2018DontGM}, \textit{XLSUM}~\cite{hasan-etal-2021-xl}, \textit{MLSUM}~\cite{scialom-etal-2020-mlsum}, \textit{Amazon Reviews} \cite{10.1145/2507157.2507163}, \textit{WikiLingua} \cite{ladhak-etal-2020-wikilingua}, \textit{COAR} \& \textit{COAH} \cite{coah}, \textit{XLM-Tweets}~\cite{barbieri-espinosaanke-camachocollados:2022:LREC}, \textit{TSATC} \cite{tsatc}, and \textit{TSD} \cite{info:doi/10.2196/14199}. 
Table \ref{tab:source_datasets} groups these datasets per domain and language.

\begin{table}[t]
\centering
\resizebox{\linewidth}{!}{%
\begin{tabular}{l|cc}
 & \multicolumn{1}{c}{\textbf{English}} & \multicolumn{1}{c}{\textbf{Spanish}} \\ \hline
\textbf{Legal} & \textit{MultiEURLEX}  & \textit{MultiEURLEX} \\
\textbf{News} & \textit{XSUM} & \textit{MLSUM} \& \textit{XLSUM} \\
\textbf{Reviews} & \textit{Amazon Reviews} & \textit{COAR} \& \textit{COAH} \\
\textbf{Tweets} & \textit{TSATC} & \textit{XLM-Tweets} \& \textit{TSD} \\
\textbf{How-to} & \textit{WikiLingua} & \textit{WikiLingua}
\end{tabular}}
\caption{\label{tab:source_datasets} Human-authored source datasets for the AuTexTification 2023 dataset.}
\end{table}

The MGT was generated from the human texts by using three \textit{BLOOM} models \cite{scao2022bloom}, \textit{BLOOM-1B7},\footnote{\url{https://tinyurl.com/bloom-1b7}} \textit{BLOOM-3B},\footnote{\url{https://tinyurl.com/bloom-3b}} and \textit{BLOOM-7B1};\footnote{\url{https://tinyurl.com/bloom7b}} as well as three \textit{GPT-3} models \cite{brown2020language,ouyang2022training}: \textit{babbage}, \textit{curie}, and \textit{text-davinci-003}, with 1b, 6.7b and 175b parameter scales respectively.
Our motivation behind using these models were fourfold: (i) both \textit{BLOOM} and \textit{GPT-3} show great capabilities in multiple languages, (ii) \textit{BLOOM} models' usage is not as restricted via licensing (as opposed to other popular models such as \textit{LLaMA} \cite{touvron2023llama} or \textit{OPT} \cite{zhang2022opt}), (iii) \textit{GPT-3} has been one of the most popular and best performing language models until recently,\footnote{\textit{GPT-3.5-turbo} and \textit{GPT-4} were not released at time of compiling our dataset.} and (iv) we aimed to cover a broad spectra of model families and scales.
While we were hoping to include \textit{BLOOM-175B} generations too, this was not possible due to the lack of public APIs.

We manually tuned the decoding parameters to obtain MGT that appears realistic through subjective evaluations carried out by two of the authors.
We found that with nucleus sampling \cite{holtzman2020curious}, using a top-p of $0.9$ and a temperature of $0.7$, the models generated texts of higher quality. 
The maximum number of completion tokens was manually selected for each domain to be similar to the median token-length of the human texts: 20 tokens for tweets, 70 for reviews, and 100 for news, legal, and how-to articles.

\subsection{Gathering process}

\begin{figure}[t]
  \centering
  \includegraphics[width=\linewidth]
  {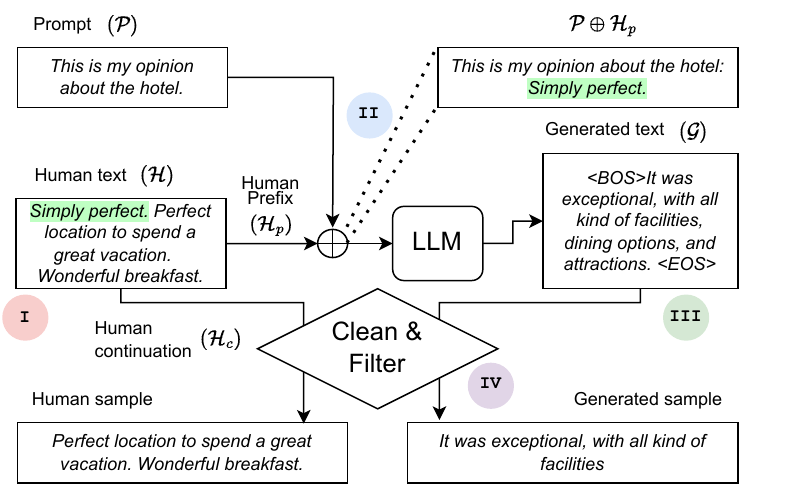}
  \caption{\label{fig:data-gathering-process}Data gathering process.}
\end{figure}

We aim to build a dataset of human and generated texts that share the same prefix. 
For instance, given a human text \say{\textit{Today it's 20 degrees. It is sunny in Valencia.}}, we could use \say{\textit{It is sunny in Valencia.}} as human text, and generate a continuation by prompting an LLM with \say{\textit{Today it's 20 degrees.}}. 
In this manner, both generated and human texts are plausible continuations of the same prefix and they can be compared fairly in terms of topics and domains. 
To build the dataset in this way, we opted for a data gathering process consisting in the steps depicted in Figure \ref{fig:data-gathering-process}, namely (i) gathering human data, (ii) preparing the inputs for LLMs, (iii) generating MGTs, and (iv) cleaning and filtering the resulting texts.

We first gather a set of human-authored texts $\mathcal{H}$ from the source datasets for each domain and language.
We manually analyze and define extraction schemes for splitting $\mathcal{H}$ into prefixes $\mathcal{H}_p$ and continuations $\mathcal{H}_c$ such that $\mathcal{H} = \mathcal{H}_p \oplus \mathcal{H}_c$. 
In some domains and source datasets, we also define prompts $\mathcal{P}$ to prevent the generation models from generating topic-inconsistent texts, e.g., guiding models to generate hotel reviews instead of car reviews when using a prefix from the \textit{COAH} dataset, made up of hotel reviews.
Afterwards, the prompts and prefixes $\mathcal{P} \oplus \mathcal{H}_p$ are fed into each LLM to obtain one resulting generation per prompt and prefix.
We refer to the set of generations as $\mathcal{G}$.
Texts from both $\mathcal{H}_c$ and $\mathcal{G}$ are fed into a text cleaning pipeline that removes duplicated spaces, multiple line breaks, and special symbols. 
Additionally, we ensure that the human continuation and generation obtained from the same prefix have roughly the same token-lengths by truncating to the minimum length of the two texts, thus removing token-length bias.
Then, we apply a set of language identification filters: \textit{langdetect},\footnote{\url{https://tinyurl.com/langdetect}} \textit{SpaCy FastLang},\footnote{\url{https://tinyurl.com/fastlang}} and \textit{fastText}~\cite{joulin2017bag}.
If one of these filters finds a text to be not in Spanish or English, the text is removed from our dataset.

To obtain the dataset for Subtask 1, we sample a subset of $\mathcal{H}_c$ labeled as \hum~and a subset of $\mathcal{G}$ labeled with \gen.
The dataset was then split into training and test sets for a cross-domain scenario: tweets, how-to articles and legal documents were included in the training set, while reviews and news data comprised the test set.
To compile the dataset for Subtask 2, we only sample texts from $\mathcal{G}$, labeling each text with the LLM's name that generated it. 
The dataset is randomly split into training and test sets following 80\%-20\% proportions. 
All the five domains are included in both training and test splits.
The released version of the dataset for Subtask 2 includes anonymized model labels to remove bias toward particular models or model families in participating submissions.

The statistics of each subtask's contents per domain, class, and language are presented in Table \ref{tab:dataset_statistics}.
In both subtasks, both languages contain similar amounts of texts, and the domains and classes are balanced in both splits.
This way we guarantee that our analysis is fair by ensuring that every dimension is balanced. 
Besides, we checked that the generated texts follow the Zipf and Heap's empirical laws, thus ensuring a high quality of the dataset.\footnote{See \url{https://tinyurl.com/overview-datasets}}

\begin{table*}
\centering
\resizebox{\linewidth}{!}{%
\begin{tabular}{cl|rrr||rrrrrrr}
                 && \multicolumn{3}{c||}{\textbf{Subtask 1 }} & \multicolumn{7}{c}{\textbf{Subtask 2 }}          \\
                 &&       &       &                             & \multicolumn{3}{c}{\textbf{BLOOM}} & \multicolumn{3}{c}{\textbf{GPT}} & \\ \hline
                 && \gen   & \hum   & $\Sigma$                    & 1b7    & 3b    & 7b1    & 1b    & 6b7    & 175b    & $\Sigma$  \\ \hline\hline
\multirow{5}{*}{\rotatebox[origin=c]{90}{\textbf{Spanish}}}&\textbf{Legal}   & 4,846  & 4,358  & 9,204                     & 640  & 665  & 712  & 919  & 942  & 919  & 4,797   \\
&\textbf{News}    & 5,514  & 5,223  & 10,737                    & 839  & 860  & 881  & 972  & 978  & 987  & 5,517   \\
&\textbf{Reviews} & 5,695  & 3,697  & 9,392                     & 952  & 962  & 935  & 945  & 941  & 947  & 5,682   \\
&\textbf{Tweets}  & 5,739  & 5,634  & 11,373                    & 967  & 965  & 965  & 928  & 930  & 964  & 5,719   \\
&\textbf{How-to}  & 5,690  & 5,795  & 11,485                    & 894  & 929  & 960  & 970  & 983  & 966  & 5,702   \\ \hline
&\textbf{Total}   & 27,484 & 24,707 & 52,191                    & 4,292 & 4,381 & 4,453 & 4,734 & 4,774 & 4,783 & 27,417  \\ \hline\hline
\multirow{5}{*}{\rotatebox[origin=c]{90}{\textbf{English}}}&\textbf{Legal}   & 5,124  & 5,244  & 10,368                    & 809  & 779  & 832  & 890  & 887  & 927  & 5,124   \\
&\textbf{News}    & 5,464  & 5,464  & 10,928                    & 747  & 854  & 906  & 983  & 984  & 984  & 5,458   \\
&\textbf{Reviews} & 5,726  & 5,178  & 10,904                    & 944  & 946  & 939  & 977  & 974  & 972  & 5,752   \\
&\textbf{Tweets}  & 5,813  & 5,884  & 11,697                    & 987  & 968  & 980  & 951  & 963  & 969  & 5,818   \\
&\textbf{How-to}  & 5,862  & 5,918  & 11,780                    & 962  & 976  & 982  & 993  & 993  & 963  & 5,869   \\ \hline
&\textbf{Total}   & 27,989 & 27,688 & 55,677                    & 4,449 & 4,523 & 4,639 & 4,794 & 4,801 & 4,815 & 28,021  \\ \hline
\end{tabular}
}
\caption{\label{tab:dataset_statistics} Number of samples per domain, class, and language in both subtasks.}
\end{table*}

\subsection{Human Assessment}
We performed a small-scale study to assess the difficulty of the Subtask 1 for human annotators. 
The study consisted in asking human annotators to classify texts as human or generated.\footnote{The annotation interface and instructions are available at \url{https://tinyurl.com/colab-annotation}} 
Five annotators were involved: four Spanish native speakers (SP) and one Italian native speaker (IT). All of them were men between the ages of 25 and 35, with C1-C2 proficiency level in English. 
From these annotators, SP1 and SP4 are familiar with generated text (they created the dataset and analysed hundreds of examples), while the others were exposed to the task for the first time.

\begin{figure}
    \centering   \includegraphics[width=\linewidth]{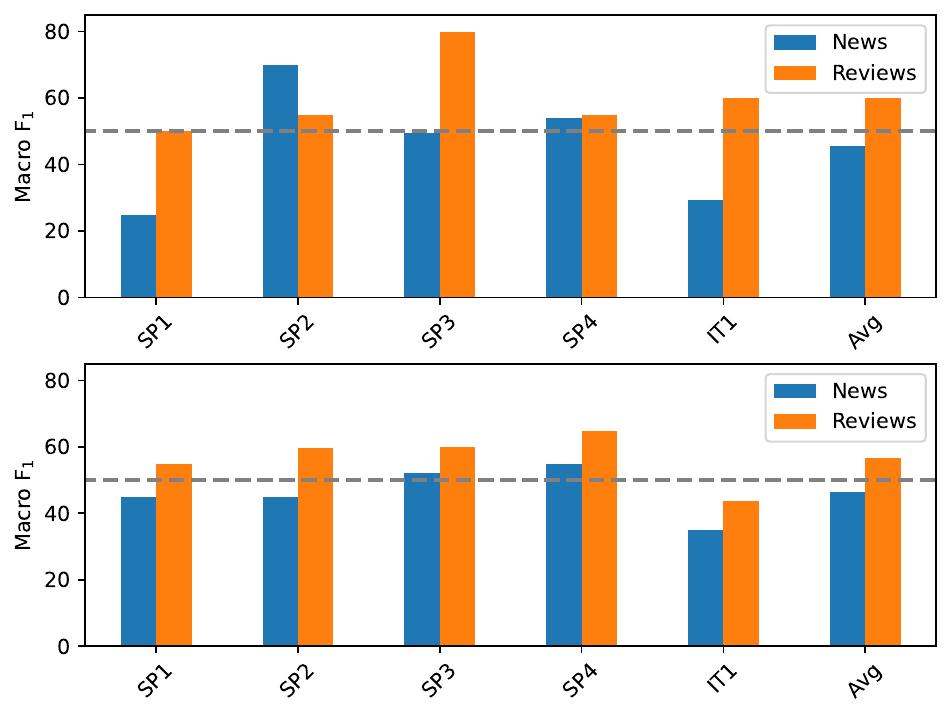}
    \caption{\label{fig:human_performance}Human performance in English (top) and Spanish (bottom). The grey dotted line is the random baseline.}
    \label{fig:enter-label}
\end{figure}

We provided the same 40 texts to each annotator, drawn from the test set of the Subtask 1 both for English and Spanish. 
The texts were balanced in terms of classes and domains: 20 texts were generated by LLMs and 20 were written by a human, half of them were news and the other half were reviews. 
The generated texts were only obtained from \textit{BLOOM} models: 6 texts from \textit{BLOOM-1b7}, 6 texts from \textit{BLOOM-3b1}, and 8 texts from \textit{BLOOM-7b1}. 
Figure \ref{fig:human_performance} shows the Macro-F$_1$ score of each annotator in each domain.

For both languages, the average annotator performance is very similar, most annotators are close to the random baseline. 
Regarding the domains, it seems more difficult for humans to distinguish between human-authored and machine-generated news rather than reviews.
Most of the annotators perform worse than the random baseline distinguishing texts from the news domain.
On the contrary, humans are typically better than the random baseline in the reviews domain, especially in English.

Language proficiency seems to play a role.
IT1 shows better performance in English than in Spanish, where he is not proficient.
Despite how SP1 and SP4 are familiar with generated texts, there seems to be no significant difference between them and other annotators.

The human annotators did not follow any systematic pattern to detect MGT.
For reviews, some mentioned that the generated reviews seemed generic, describing many general aspects with short sentences.
In contrast, human reviews focused on few and more concrete aspects.

\section{Systems and Results}
In this section, we briefly introduce the participants' systems, describe the baselines and evaluation metrics, and study the results of the shared task.

\subsection{Submitted Approaches}

The AuTexTification shared task received submissions from 36 teams, belonging to 30 different institutions and 18 different countries. 
All teams participated in the English track of Subtask 1, with 23 teams also taking part in the Spanish track. 
For Subtask 2, 19 teams participated in the English track and 14 in the Spanish track. 
Teams were allowed to submit a maximum of 3 runs per subtask and language. 
Overall, AuTexTification received a total of 175 runs, comprising 71 for the English track of Subtask 1, 47 for the Spanish track, 33 for the English track of Subtask 2, and 24 for the Spanish track. 
Outside of the competition scope, the AuTexTification datasets have been used in NLP courses within academic institutions. We are aware of at least 3 institutions,\footnote{Universitat Politècnica de València, Aix-Marseille Université, and IMT Atlantique.} with 17 participating teams and 58 runs.

Following the trend in the Natural Language Processing (NLP) field, most teams relied on pre-trained Transformer \cite{vaswani2017attention} models. 
The most used ones were BERT-based models \cite{devlin-etal-2019-bert} like \textit{RoBERTa} \cite{liu2019roberta} and \textit{DeBERTa} \cite{he2021debertav3}. Also, domain-specific and multilingual variants of \textit{BERT} were frequent, including \textit{XLM-RoBERTa} \cite{conneau2020unsupervised}, \textit{RemBERT} \cite{chungrethinking}, and \textit{Twhin-BERT} \cite{zhang2022twhin}. A smaller set of participants included generative models in their systems such as \textit{GPT-2} \cite{radford2019language}, \textit{Grover} \cite{NEURIPS2019_3e9f0fc9}, and \textit{OPT} \cite{zhang2022opt}.

Most of the best performing approaches used ensembles of pre-trained models, as well as combinations of lexical, stylometric or statistical features.
In some cases, participants fine-tuned their models using auto-train procedures and performed hyper-parameter tuning. 
Some teams also included \textit{Convolutional Neural Networks} \cite{lecun1989handwritten} or \textit{Long Short Term Memory (LSTM) Networks} \cite{hochreiter1997long} as part of their systems.
Traditional machine learning models like \textit{Logistic Regression} and \textit{Support Vector Machines (SVM)} \cite{cortes1995support} were also frequent among the participants.
However, these approaches generally performed worse than Transformer-based approaches.

There was also a great diversity in terms of features. 
Probabilistic token-level features from generative language models seem to play an important role in the best performing approaches. 
Most participants used contextual representations from pre-trained models, either as features, or through end-to-end fine-tuning.
Linguistic features including lexical, structural, and discourse features were also frequent. 
Among the most common linguistic features, we observed bag of word/char n-grams, counts of personal pronouns, stop-words, punctuations, and POS tags.
Some participants also incorporated linguistic and factual knowledge directly in their models.
Among these, we found the inclusion of syntactic dependencies in pre-trained models through contrastive learning, Wikipedia fact-checking, and native language identification.

The best ranked systems for each subtask ranged from complex ensembles of many different models and features, to single generative models fine-tuned for the task. 
In Subtask 1, both for English and Spanish, the best system was proposed by \textit{TALN-UPF}~\cite{taln_upf}.
This system relied on a bidirectional \textit{LSTM} \cite{schuster1997bidirectional} model trained with a combination of probabilistic token-level features from different \textit{GPT-2} versions, linguistic token-level features such as word-frequencies or grammar errors, and text representations from pre-trained encoders. 
Besides, \textit{TALN-UPF} was the only team that considered a cross-domain evaluation in the validation step, by performing cross-validation over topically-split data after inferring the topics using \textit{Latent Dirichlet Allocation} \cite{blei2003latent}. 
In the Spanish track, the \textit{TALN-UPF} system performed similar to the \textit{Lingüística\_UCM} system~\cite{linguistica_ucm}, consisting of an \textit{SVM} trained with a set of morphological, lexical, and discourse features selected according linguistic expertise and human analysis.

In Subtask 2, both for English and Spanish, the three runs of the \textit{Drocks} team~\cite{drocks} were the highest ranked ones. 
These systems were ensembles of five different Transformer-based classifiers fine-tuned on the task. 
The best ensembles differed for each language. 
For English, the best ensemble was an \textit{Error-Correcting Output Codes}~\cite{dietterich1994solving} model trained using the concatenation of the classification probabilities as features. 
For Spanish, the best ensemble was implemented with an \textit{SVM} using the average of the classification probabilities as features.

\subsection{Baselines}
We consider several baselines for each subtask and language.
Namely, we include a random baseline (\textit{Random}), zero-shot (\textit{SB-ZS}) and few-shot (\textit{SB-FS}) approaches based on text and label embedding similarities, a bag-of-words encoding with logistic regression (\textit{BOW+LR}), Low Dimensional Semantic Embeddings (\textit{LDSE}), and fine-tuned language specific transformers (Transformer), \textit{DeBERTaV3} \cite{he2021debertav3}\footnote{\url{https://tinyurl.com/debertav3}} for English and \textit{RoBERTa-BNE} \cite{gutierrez2022maria}\footnote{\url{https://tinyurl.com/robertabne}} for Spanish.
These baselines consist in the following:

\paragraph{Random.} The random baseline assuming class balance.
Defined as $\frac{1}{C}$ where $C$ is the number of classes.

\paragraph{SB-ZS and SB-FS.} Zero-shot and Few-Shot Symanto Brain API,\footnote{\label{foot:sb}\url{https://www.symanto.com/nlp-tools/symanto-brain/}} a \copyright Symanto solution optimized for highly efficient and scalable state-of-the-art zero-shot and few-shot classification \cite{mueller2022few}.  
We verbalize labels for Subtask 1,\footnote{\hum: \say{\textit{This text has been written by a human.}}\\ \gen: \say{\textit{This text has been automatically generated by a bot.}}} but not for Subtask 2 given the anonymity of the classes. 
For \textit{SB-FS} we use 1024 shots. 

\paragraph{BOW+LR.} We encode the texts with bag of n-grams, using the top 5K word $n$-grams, $n\in\set{1, 2}$ and character $n$-grams, $n\in\set{2,\ldots,6}$ following \cite{pizarro2019using}.
We train a \textit{Logistic Regression} model offered by scikit-learn \cite{scikit-learn} with default parameters on z-score normalized and concatenated features. 

\paragraph{LDSE.} We represent texts on the basis of the probability distribution of occurrence of their tokens in the different classes with \textit{LDSE}~\cite{rangel2018low}.
We train an \textit{SVM} classifier provided by scikit-learn \cite{scikit-learn} with default parameters.

\paragraph{Transformer.} We use the HuggingFace ecosystem \cite{wolf2020transformers} to fine-tune a pre-trained Transformer with a randomly initialized classification head for 5 epochs and default hyperparameters. 
We use a batch size of 32 texts for \textit{DeBERTaV3} and a batch size of 64 for \textit{RoBERTa-BNE}.

\subsection{Evaluation}
The submissions for both subtasks are evaluated with the Macro-F$_1$ score. 
Statistical significance is computed through bootstrapping with replacement at a confidence level of $\alpha=0.95$ with 1,000 resamples.

\subsection{Subtask 1: MGT Detection}

For the MGT detection subtask, we received 71 submissions from 36 different teams in English, and 47 submissions from 23 teams in Spanish.
Tables \ref{tab:subtask_1_ranking_en} and \ref{tab:subtask_1_ranking_es} show the top-3 performing teams, the weakest team, as well as the first team that beats each baseline, both for English and Spanish.

\begin{table}[t]
\centering
\scriptsize
\setlength\tabcolsep{4.5pt}
\begin{tabular}{lrrr}
\hline
\multicolumn{1}{c}{\textbf{Rank}} & \multicolumn{1}{c}{\textbf{Team}} & \multicolumn{1}{c}{\textbf{Run}} & \multicolumn{1}{c}{\textbf{Macro-F$_1$}} \\ \hline
1  & TALN-UPF       & HB\_plus & \textbf{80.91} \\
2  & TALN-UPF       & HB       & 74.16 \\
3  & CIC-IPN-CsCog  & run2     & 74.13 \\
22  & turquoise\_titans  & run1     & 65.79 \\
23 & {\color{orange}BOW+LR}   & {\color{orange}baseline} & 65.78 \\
33 & turing\_testers & run3 & 60.64 \\
34 & {\color{orange}LDSE}   & {\color{orange}baseline} & 60.35 \\
37 & OD-21          & run3     & 59.49 \\
38 & {\color{orange}SB-FS}  & {\color{orange}baseline} & 59.44 \\
51 & swissnlp\_team & run2     & 57.20  \\
52 & {\color{orange}Transformer}     & {\color{orange}baseline} & 57.10  \\
69 & UMZ            & run1     & 50.18 \\
70 & {\color{orange}Random}         & {\color{orange}baseline} & 50.00    \\
74 & {\color{orange}SB-ZS} & {\color{orange}baseline} & 43.47 \\
77 & UAEMex         & run1     & 33.87 \\ \hline
\end{tabular}
\caption{\label{tab:subtask_1_ranking_en}Ranking of Subtask 1 (English).}
\end{table}

\begin{table}[t]
\centering
\scriptsize
\setlength\tabcolsep{5pt}
\begin{tabular}{lrrr}
\hline
\multicolumn{1}{c}{\textbf{Rank}} & \multicolumn{1}{c}{\textbf{Team}} & \multicolumn{1}{c}{\textbf{Run}} & \multicolumn{1}{c}{\textbf{Macro-F$_1$}} \\ \hline
1  & TALN-UPF          & HB\_plus & \textbf{70.77} \\
2  & Ling\_UCM & run1     & 70.60  \\
3  & {\color{orange}Transformer}       & {\color{orange}baseline} & 68.52 \\
20  & {GLPSI}       & {run3} & 63.90 \\
21 & {\color{orange}LDSE} & {\color{orange}baseline} & 63.58 \\
25 & turing\_testers   & run1     & 62.77 \\
26 & {\color{orange}BOW+LR}      & {\color{orange}baseline} & 62.40  \\
39 & bucharest         & run2     & 56.49 \\
40 & {\color{orange}SB-FS}     & {\color{orange}baseline} & 56.05 \\
46 & ANLP              & run1     & 51.38 \\
47 & {\color{orange}Random}            & {\color{orange}baseline} & 50.00    \\
50 & UAEMex            & run3     & 35.17 \\
51 & {\color{orange}SB-ZS}    & {\color{orange}baseline} & 34.58 \\
53 & LKE\_BUAP         & run3     & 31.60  \\ \hline
\end{tabular}
\caption{\label{tab:subtask_1_ranking_es}Ranking of Subtask 1 (Spanish).}
\end{table}

\begin{figure}
    \centering
    \includegraphics[width=\linewidth]{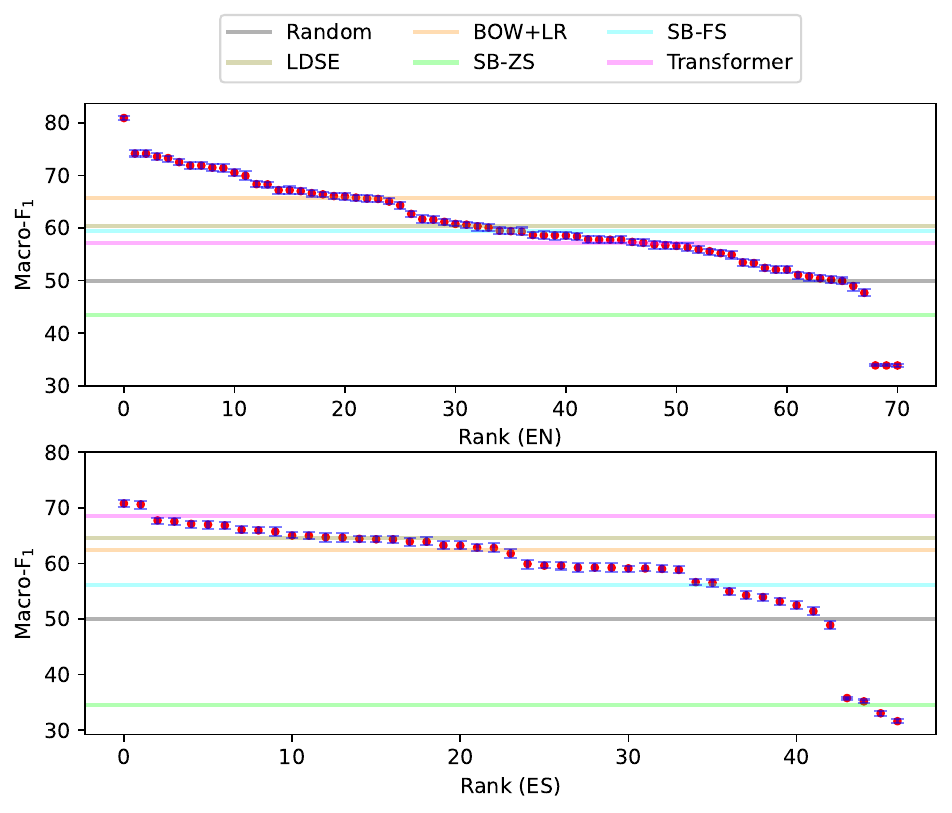}
    \caption{\label{fig:subtask_1_errorbar_mf1}Rank-ordered Macro-F$_1$ with error bars for Subtask 1 in English (top) and Spanish (bottom). Colored lines are baselines.}
\end{figure}

The best system was proposed by the \textit{TALN-UPF} team, with 80.91 and 70.77 Macro-F$_1$ scores in English and Spanish. 
In English, the best team is significantly better than the second-best ranked team. 
However, in Spanish there are no significant differences between the two best teams and the best baseline. 
In Figure \ref{fig:subtask_1_errorbar_mf1}, we illustrate the rank-ordered Macro-F$_1$ scores for all the teams in both languages.

\begin{figure*}[t]
    \centering
    \begin{subfigure}{.33\linewidth}
        \centering \includegraphics[width=\linewidth]{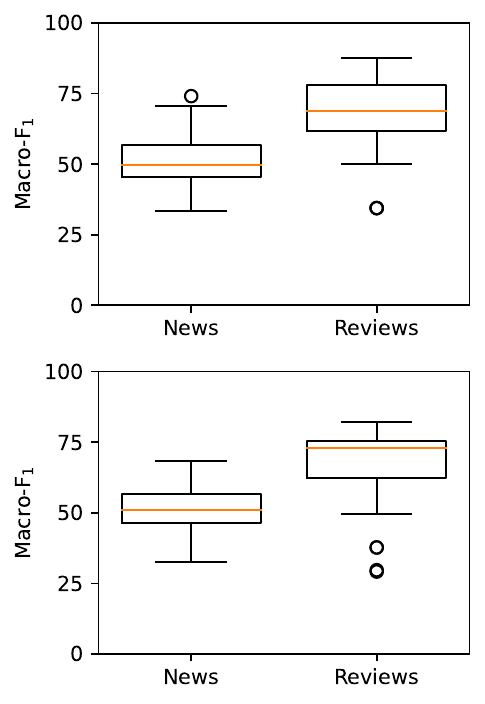}
        \caption{\label{fig:subtask_1_domainwise_boxplots}Per domain Macro-F$_1$.}
    \end{subfigure}%
    \begin{subfigure}{.33\linewidth}
        \centering \includegraphics[width=\linewidth]{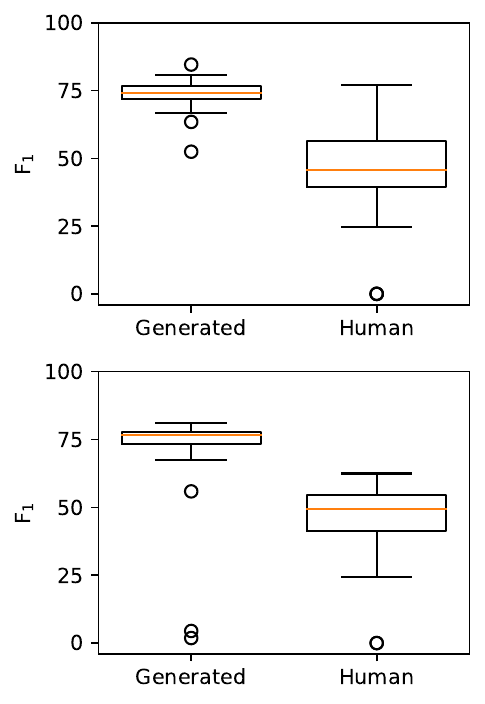}
    \caption{\label{fig:subtask_1_classwise_boxplots}Per class F$_1$.}
    \end{subfigure}%
    \begin{subfigure}{.34\linewidth}
        \begin{minipage}{\linewidth}
            \centering
            \includegraphics[width=\linewidth, height=3.75cm]{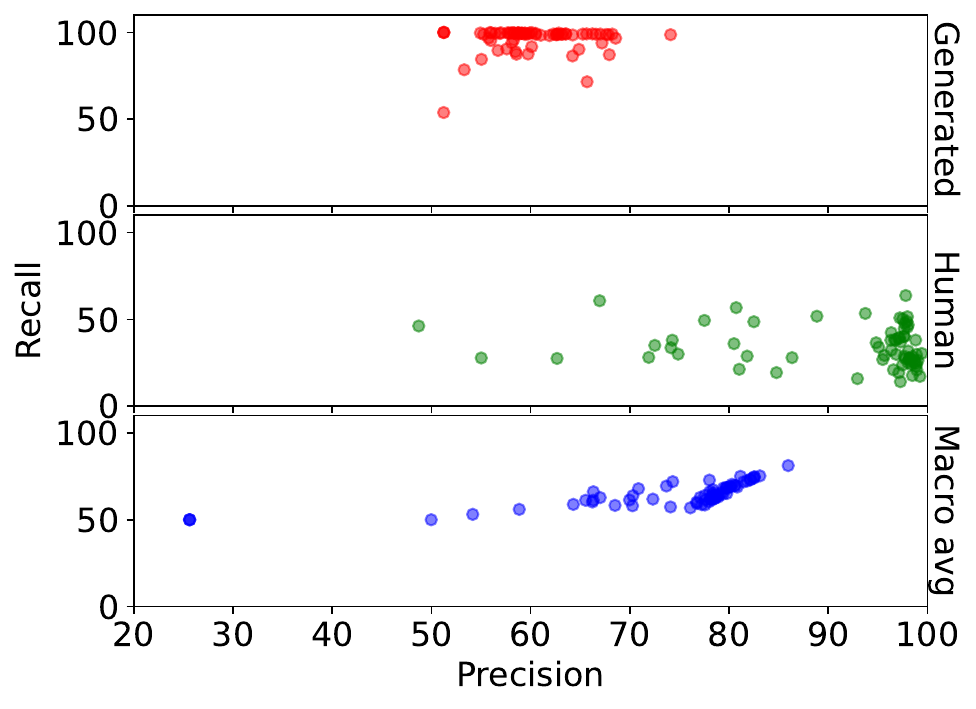}
        \end{minipage}
        \begin{minipage}{\linewidth}
            \centering
            \includegraphics[width=\linewidth, height=3.75cm]{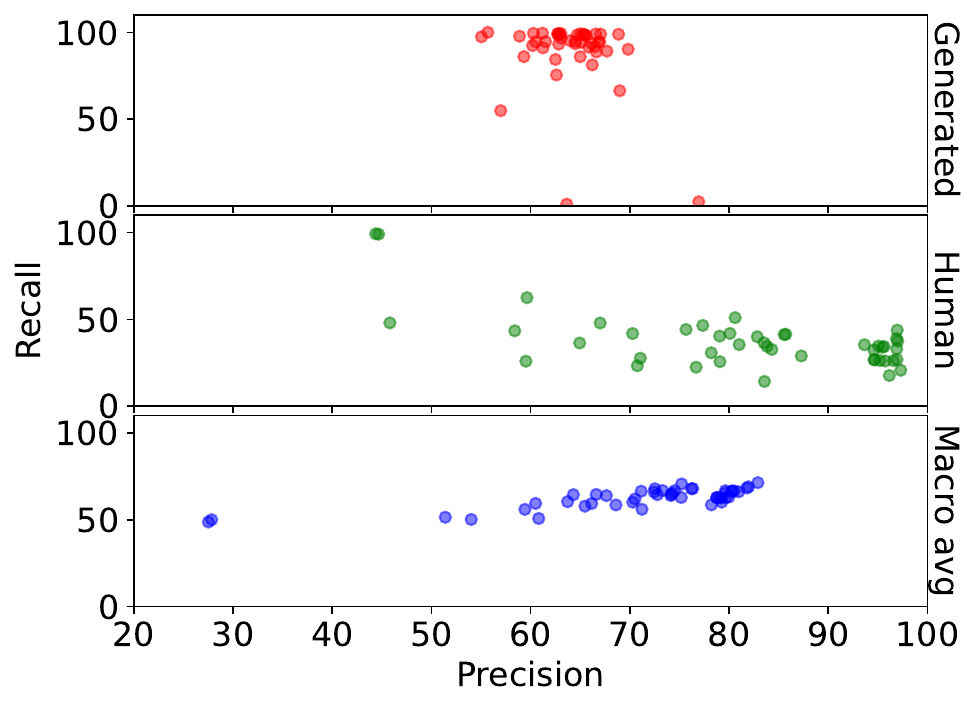}
        \end{minipage}
    \caption{\label{fig:subtask_1_precision_recall}Per class precision-recall plots.}
    \end{subfigure}
    \caption{\label{fig:subtask_1_boxplots_prscatters}Fine-grained plots for Subtask 1 in English (top) and Spanish (bottom).}
\end{figure*}

Many teams surpassed the best baseline in English by large margins, whereas for Spanish only two teams were able to outperform it with small differences in Macro-F$_1$. 
Moreover, the performance of the top-11 ranked teams in English is higher than the performance of the best team in Spanish. 
This could suggest that detecting MGT and generalizing to new domains is easier in English than in Spanish, either due to language idiosyncrasies or because of the larger availability and quality of English NLP models. 
For both languages, we observe a linear relationship between the rank-ordered Macro-F$_1$ scores, with a small set of outliers in both tails. 
This hints that, even though the resulting Macro-F$_1$ scores in each language are in different ranges, there is similar variability and difficulty in both languages. 
The teams' systems cover almost the entire Macro-F$_1$ range in both languages, and, in many cases, they are very similar (same Transformer-based models, similar linguistic features, etc.). Therefore, one has to be careful when developing a MGT detector, small changes could lead to large improvements or declines.

We also include fine-grained results per-domain and per-class in Figure \ref{fig:subtask_1_boxplots_prscatters}. 
When observing the domain-wise Macro-F$_1$ scores in Figure \ref{fig:subtask_1_domainwise_boxplots}, we find that the systems generalized better to reviews than to news, with a mean Macro-F$_1$ below the random baseline for the latter. 
Furthermore, both domains show long-tailed distributions, revealing the variability in generalization capabilities of the systems. 
Concerning class-wise F$_1$ scores in Figure \ref{fig:subtask_1_classwise_boxplots}, we find that the systems are better at classifying generated text, and there is lower dispersion among the systems' F$_1$ scores for this class than for the human class.
From the precision-recall distributions depicted in Figure \ref{fig:subtask_1_precision_recall}, we observe that systems are more biased towards predicting text to be generated (high recall), often doing so incorrectly (low precision). 
We observe the opposite for human texts, few predictions (low recall) that are mostly correct (high precision). 
All the conclusions above hold for both languages.

For the sake of completeness, we refer the reader to the AuTexTification repository,\footnote{\label{foot:overview_repo}\url{https://tinyurl.com/overview-results}} which includes additional plots, the most difficult and easiest examples for the systems, complete rankings including submissions outside the competition, etc.

\subsection{Subtask 2: MGT Attribution}

For the MGT Attribution subtask we received 33 submissions from 19 different teams in English, and 24 submissions from 14 teams in Spanish. Tables \ref{tab:subtask_2_ranking_en} and \ref{tab:subtask_2_ranking_es} show the top-3 performing teams, the weakest team, as well as the first team that beats each baseline, both for English and Spanish.

\begin{figure}[t]
    \centering
    \includegraphics[width=\linewidth]{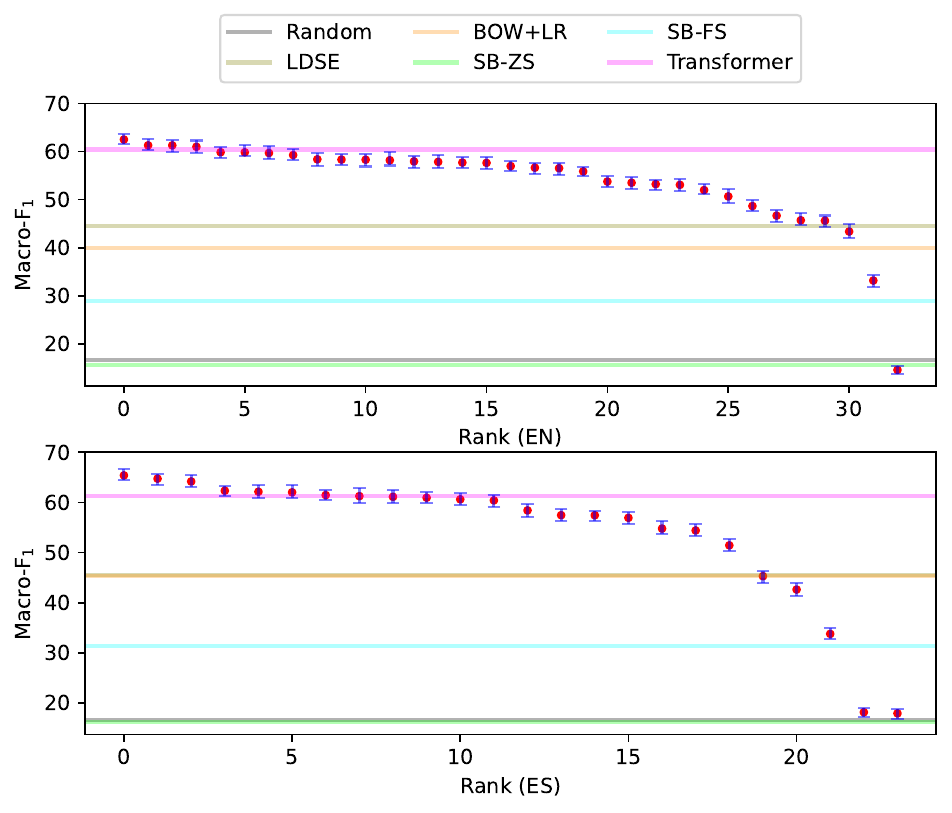}
    \caption{\label{fig:subtask_2_errorbar_mf1}Rank-ordered Macro-F$_1$ for Subtask 2 in English (top) and Spanish (bottom). Colored lines are baselines.}
\end{figure}

\begin{figure*}[t]
  \begin{subfigure}[t]{0.33\linewidth}
    \includegraphics[width=\linewidth, height=21em]{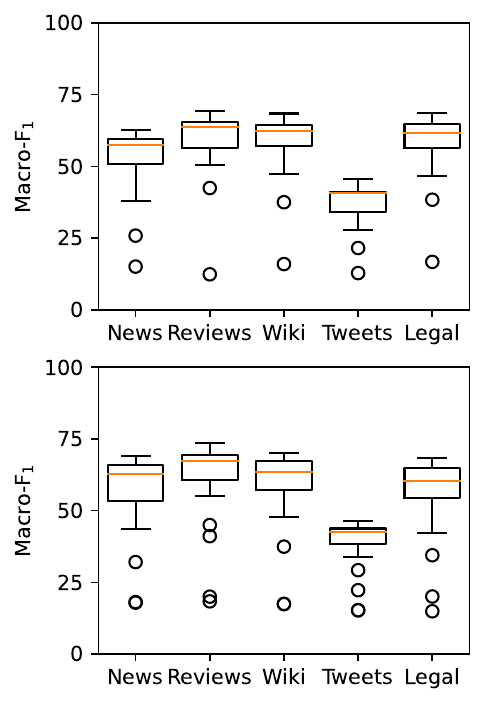}
    \caption{\label{fig:subtask_2_domainwise_boxplots}Per domain Macro-F$_1$.}
  \end{subfigure}\hfill
  \begin{subfigure}[t]{0.33\linewidth}
    \includegraphics[width=\linewidth, height=21em]{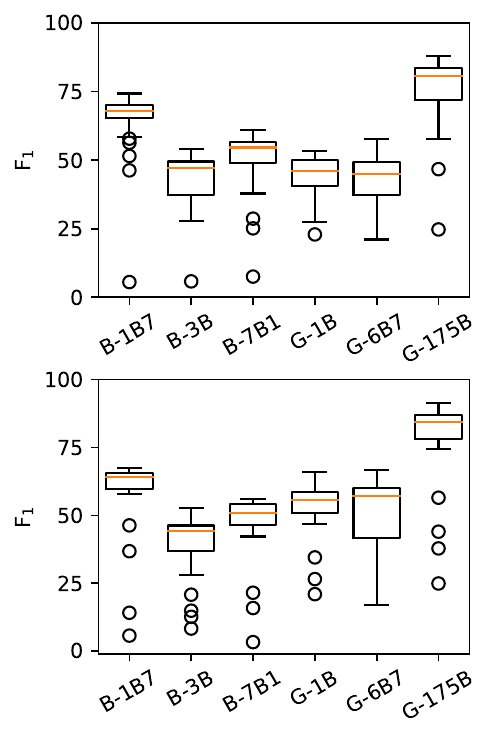}
    \caption{\label{fig:subtask_2_classwise_boxplots}Per class F$_1$.}
  \end{subfigure}
  \begin{subfigure}[t]{0.33\linewidth}
    \includegraphics[width=\linewidth, height=21em]{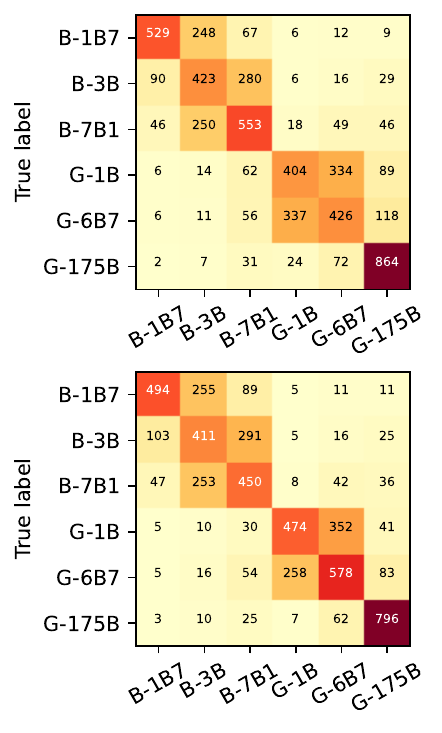}
    \caption{\label{fig:subtask_2_confusions}Median confusion matrices.}
  \end{subfigure}
  \caption{\label{fig:subtask_2_boxplots}Fine-grained plots for Subtask 2 in English (top) and Spanish (bottom). B- prefix denotes \textit{BLOOM} models and G- prefix denotes \textit{GPT} models.}
\end{figure*}

The best system was submitted by team \textit{Drocks}, obtaining 62.5 and 65.37 Macro-F$_1$ scores for English and Spanish, respectively.
This is in contrast to the best scores of Subtask 1 nearing 80 and 70 Macro-F$_1$, showing that in-domain MGT attribution is more difficult than out-of-domain MGT detection.
In this subtask, teams did not deviate significantly from the baselines, and for both languages the relative ranking of baselines remained the same, as opposed to Subtask 1.
Rank ordered Macro-F$_1$ scores for both languages are presented in Figure \ref{fig:subtask_2_errorbar_mf1}.
Few teams were able to surpass the best baselines, with most submissions performing between the top-2 baseline scores.
Similarly to Subtask 1, we observe a linear relationship between rank and Macro-F$_1$ with outliers in the right tail, meaning that there is variability and difficulty in attributing MGT irrespective of language.
However, teams cover a smaller range of Macro-F$_1$ scores than in Subtask 1, suggesting there is less variability when attributing MGT than detecting it.
In contrast to Subtask 1, teams generally obtained better Macro-F$_1$ scores in Spanish than English, but the differences were marginal, which could be because of the randomness of the learning procedures or due to a smaller number of participants in Subtask 2.
Generally, MGT attribution appears promising but limited, suggesting the need for further research into new approaches or framings of the problem.
\begin{table}[t]
\centering
\scriptsize
\setlength\tabcolsep{4.5pt}
\begin{tabular}{lrrr}
\hline
\multicolumn{1}{c}{\textbf{Rank}} & \multicolumn{1}{c}{\textbf{Team}} & \multicolumn{1}{c}{\textbf{Run}} & \multicolumn{1}{c}{\textbf{Macro-F$_1$}} \\ \hline
1  & Drocks       & run3 & \textbf{62.50} \\
2  & Drocks       & run1       & 61.29 \\
3  & Drocks  & run2     & 61.27 \\
4  & ViDa  & run1     & 60.99 \\
5 & {\color{orange}Transformer}   & {\color{orange}baseline} & 60.42 \\
31 & LKE\_BUAP & run1 & 45.62 \\
32 & {\color{orange}LDSE}   & {\color{orange}baseline} & 44.56 \\
33 & turquoise\_titans          & run2     & 43.37 \\
34 & {\color{orange}BOW+LR}  & {\color{orange}baseline} & 39.98 \\
35 & UAEMex & run2     & 33.19  \\
36 & {\color{orange}SB-FS}     & {\color{orange}baseline} & 28.94  \\
37 & {\color{orange}Random}         & {\color{orange}baseline} & 16.66    \\
38 & {\color{orange}SB-ZS} & {\color{orange}baseline} & 15.70 \\
39 & ANLP         & run1     & 14.61 \\ \hline
\end{tabular}
\caption{\label{tab:subtask_2_ranking_en}Ranking Subtask 2 (English).}
\end{table}
Fine-grained per-domain and per-class results for Subtask 2 are presented in Figure \ref{fig:subtask_2_boxplots}.
Per-domain results (Figure \ref{fig:subtask_2_domainwise_boxplots}) show that attribution of generated tweets is much more difficult than the remaining domains.
For tweets, systems are unable to reach 50\% Macro-F$_1$, while for the other domains they surpass it by a large margin.
We additionally find many outliers toward lower scores, indicating the difficulty of the task.
Finally, most domains have similar distributions centered around different medians, meaning that the variability of participating systems is maintained through all five domains.
We also present per-class results in Figure \ref{fig:subtask_2_classwise_boxplots}, where we find that it is easier to attribute generated text to \textit{BLOOM-1B7} and \textit{text-davinici-003}.
Moreover we observe large variability for \textit{curie}, while the other classes have  narrower distributions.

Additionally, we computed overall confusion matrices by taking the median at each position of the confusion matrix from all the participant's systems. Figure \ref{fig:subtask_2_confusions} shows the results for English and Spanish. In both languages, the largest confusions are across models within the same families, 
suggesting that it is easier to distinguish generation models of different families. Besides, \textit{text-davinci-003} is the model with less number of confusions, being different enough to be easily distinguished from the other models.

Once again, we refer to the AuTexTification repository\footref{foot:overview_repo} for additional plots, results and analyses.

\begin{table}[t]
\centering
\scriptsize
\setlength\tabcolsep{4.5pt}
\begin{tabular}{lrrr}
\hline
\multicolumn{1}{c}{\textbf{Rank}} & \multicolumn{1}{c}{\textbf{Team}} & \multicolumn{1}{c}{\textbf{Run}} & \multicolumn{1}{c}{\textbf{Macro-F$_1$}} \\ \hline
1  & Drocks       & run2 & \textbf{65.37} \\
2  & Drocks       & run3       & 64.72 \\
3  & Drocks  & run1     & 64.17 \\
7  & TALN-UPF  & Hybrid\_plus     & 61.45 \\
8 & {\color{orange}Transformer}   & {\color{orange}baseline} & 61.34 \\
20 & iimasPLN & run1 & 51.43 \\
21 & {\color{orange}LDSE}   & {\color{orange}baseline} & 45.46 \\
22 & {\color{orange}BOW+LR}  & {\color{orange}baseline} & 45.31 \\
25 & UAEMex & run2     & 33.78  \\
26 & {\color{orange}SB-FS}     & {\color{orange}baseline} & 31.38  \\
28 & ANLP     & run1 & 17.93  \\
29 & {\color{orange}Random}         & {\color{orange}baseline} & 16.66    \\
30 & {\color{orange}SB-ZS} & {\color{orange}baseline} & 16.23 \\ \hline
\end{tabular}
\caption{\label{tab:subtask_2_ranking_es}Ranking Subtask 2 (Spanish).}
\end{table}

\section{Conclusions and Future Work}
This paper describes the AuTexTification shared task at IberLEF 2023, which aimed to study the automatic detection of MGT in cross-domain scenarios and MGT attribution to specific generation models, across five domains and two languages.
The AuTexTification dataset was comprised of around 160,000 texts collected through an automatic data gathering process which can be easily extended to new domains and languages.
The task received a significant amount of participation: 175 runs from 36 teams, belonging to 30 different institutions and 18 different countries, thus showing the overall interest of the community in addressing MGT detection and attribution. Moreover, other 17 teams submitted 58 runs although after the deadline, for a total of 233 runs by 53 teams.

The participating systems relied on a wide variety of approaches, with a strong trend towards the use of pre-trained Transformer models.
Ensembles of pre-trained models and combinations of probabilistic, lexical, and stylometric features led to the best performing systems in both subtasks. 
The results suggest that cross-domain MGT detection is easier in English than in Spanish, and that MGT attribution is generally more challenging than MGT detection. 
While MGT attribution appears promising, the small gap between the participant's systems and the baselines encourage further research.
Overall, the results suggest that MGT detection and attribution remain challenging tasks and there is potential for further progress.

As future work, we hope to expand the AuTexTification dataset to include more languages, domains, generation models and decoding strategies, to encourage the development of more robust and generalizable systems. 
Furthermore, it would be valuable to explore alternative formulations of MGT attribution, as fine-grained attribution remains a challenging task. 

\section*{Acknowledgements}
We would like to thank Guillermo Pérez-Torró, Ian Borrego Obrador, and Angelo Basile for their precious help participating in the human assessment, and Mara Chinea Rios for developing a custom implementation of the LDSE baseline. 

The work from Symanto has been partially funded by the Pro$^2$Haters - Proactive Profiling of Hate Speech Spreaders (CDTi IDI-20210776), the XAI-DisInfodemics: eXplainable AI for disinformation and conspiracy detection during infodemics (MICIN PLEC2021-007681), the OBULEX - \textit{OBservatorio del Uso de Lenguage sEXista en la red} (IVACE IMINOD/2022/106), and the ANDHI - ANomalous Diffusion of Harmful Information (CPP2021-008994) R\&D grants.
The work of Areg Mikael Sarvazyan has been partially developed with the support of valgrAI - Valencian Graduate School and Research Network of Artificial Intelligence and the Generalitat Valenciana, and co-founded by the European Union. 
The research at the Universitat Politècnica de València was framed under the FairTransNLP research project, Grant
PID2021-124361OB-C31 funded by MCIN/AEI/10.13039/501100011033 and by ERDF, EU A way of making Europe.

\bibliographystyle{fullname}
\bibliography{main}

\end{document}